\documentclass[runningheads]{llncs}
\usepackage{amsfonts}
\usepackage{graphicx}
\usepackage{subcaption}
\usepackage{amsmath}
\usepackage{nicefrac}
\usepackage{multirow}
\usepackage{hyperref}
\begin{document}
\title{Sparse Bayesian Networks: Efficient Uncertainty Quantification in Medical Image Analysis}

\titlerunning{Sparse Bayesian Networks for Uncertainty in Medical Image Analysis}
\author{Zeinab Abboud\inst{1} \and 
Herve Lombaert\inst{1} \and
Samuel Kadoury\inst{1}}
% Third Author\inst{3}\orcidID{2222--3333-4444-5555}
%
\authorrunning{Z. Abboud et al.}
\institute{Polytechnique Montreal}
\maketitle              % typeset the header of the contribution
\begin{abstract}
Efficiently quantifying predictive uncertainty in medical images remains a challenge. While Bayesian neural networks (BNN) offer predictive uncertainty, they require substantial computational resources to train. Although Bayesian approximations such as ensembles have shown promise, they still suffer from high training and inference costs. Existing approaches mainly address the costs of BNN inference post-training, with little focus on improving training efficiency and reducing parameter complexity. This study introduces a training procedure for a sparse (partial) Bayesian network. Our method selectively assigns a subset of parameters as Bayesian by assessing their deterministic saliency through gradient sensitivity analysis. The resulting network combines deterministic and Bayesian parameters, exploiting the advantages of both representations to achieve high task-specific performance and minimize predictive uncertainty. Demonstrated on multi-label ChestMNIST for classification and ISIC, LIDC-IDRI for segmentation, our approach achieves competitive performance and predictive uncertainty estimation by reducing Bayesian parameters by over 95\%, significantly reducing computational expenses compared to fully Bayesian and ensemble methods.

\end{abstract}
\section{Introduction}
\begin{figure}[!t]
\begin{center}
    \begin{subfigure}{0.22\columnwidth}
    \centering
        \includegraphics[width=0.7\linewidth]{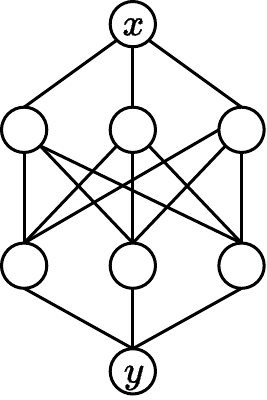}
        \caption{Deterministic}\label{fig:deterministic}
    \end{subfigure}
    \hfill
    \begin{subfigure}{0.22\columnwidth}
    \centering
        \includegraphics[width=0.7\linewidth]{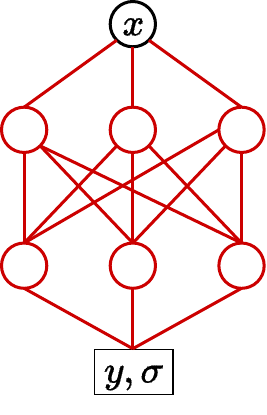}
        \caption{Bayes}\label{fig:bayes}
    \end{subfigure}
    \hfill
    \begin{subfigure}{0.22\columnwidth}
    \centering
        \includegraphics[width=0.7\linewidth]{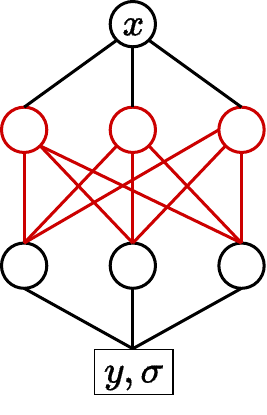}
        \caption{Layer-Partial}\label{fig:layer-wise}
    \end{subfigure}
    \hfill
    \begin{subfigure}{0.22\columnwidth}
    \centering
        \includegraphics[width=0.7\linewidth]{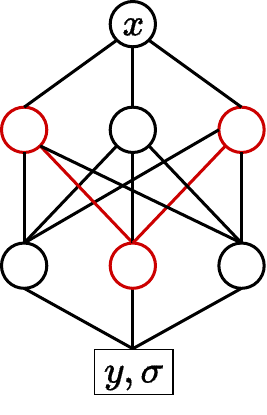}
        \caption{Sparse-Partial} \label{fig:sparse-wise}
    \end{subfigure}
    \caption{Various model implementations of deterministic (a), Bayesian (b), partial Bayesian (c, d), where black connections are deterministic, and red are probabilistic (Bayesian). Partial Bayesian models can be implemented in two distinct approaches, (c) where a single or multiple layers can be set as Bayesian, or sparse approach (d) where a selected number of connections can be Bayesian. (b~$\rightarrow$ d arranged from highest to lowest number of Bayesian parameters).}
    \label{fig:diff_models}
\end{center}
% \vskip -0.2in
\end{figure}

Unlocking the full potential of deep learning (DL) diagnostic systems in medical imaging crucially depends on precision in gauging predictive uncertainty. Without a firm grasp on uncertainty, effectively quantifying and conveying the risks tied to model predictions becomes difficult for clinicians \cite{mccrindle2021radiology}. As opposed to qualitative measures of uncertainty, such as saliency maps \cite{reyes2020interpretability}, predictive uncertainty is a quantitative measure of confidence or lack thereof in model prediction \cite{mccrindle2021radiology}. DL algorithms often fail to estimate model uncertainty, leading to unreliable predictions with overconfident false classifications \cite{guo17calibration,gal2016thesis}. This issue arises as DL models commonly assign poorly calibrated probabilities, posing risks in interpretation and decision-making \cite{guo17calibration,gal2016thesis}. BNNs represent parameters as random variables, characterized by distributions rather than single point estimates (Figure \ref{fig:bayes}). BNNs, therefore, provide average predictions and uncertainty estimates by sampling from their parameter distributions. The major limitations of BNNs stem from their significant computational costs and noisy loss landscape associated with increased training complexity and high parameter count \cite{jospin2022hands}. 

Recent attempts have been made to reduce the computational costs associated with Bayesian \textit{inference} by post-hoc processing methods \cite{Jia2021BNNinference,subedarquantization,sharma2021bayesian}; however, they do not address the costs associated with training. Few studies have utilized a strategy of conducting Bayesian inference on the output layer of a given neural network \cite{azizzadenesheli2018efficient,jospin2022hands} as a means of reducing the computational overhead (see Figure \ref{fig:layer-wise}). While previous research has explored strategies for optimizing Bayesian layer placement and justifying its selection \cite{zeng2018relevance,prabhudesai2023lowering}, there remains a lack of investigation into further decreasing the parameter complexity of Bayesian training and inference. While there exist other Bayesian approximation methods, such as Monte-Carlo (MC) dropout \cite{gal2016dropout} and deep ensembles \cite{lakshminarayanan2017simple}, they have their limitations. MC-dropout exhibits overconfidence, posing risks in safety-critical applications, while deep ensembles incur high computational costs due to training and storing multiple models for inference. Deep ensembles are especially impractical with models with large parameter counts, which is common in medical imaging.

In contrast to Bayesian models, there have been successful attempts to utilize sparsity for reducing model parameter complexities in deterministic models \cite{NIPS2016_guoSparsePruning,NIPS1989_optimalBrainDamage,dey2019pre-defined,evci2020rigging}. 
Parameter sparsity involves pruning many connections within a neural network to accelerate model convergence and reduce computational overhead. Early studies involve reducing network size by pruning the weights and re-training through saliency analysis \cite{NIPS1989_optimalBrainDamage}. LeCun et al. \cite{NIPS1989_optimalBrainDamage} show that reducing the number of parameters by more than 30\% resulted in no impact on the model performance. Other methods have used second-order derivatives, Hessian-based analysis \cite{NIPS1988_trimmingthefat,NIPS1992_brainSurgeon} to prune weights; however, such approaches are impractical with larger and more complex modern networks. More recent approaches introduce parameter sparsity through pre-defined randomized sparsity \cite{dey2019pre-defined}, structured sparsity \cite{wen16_structuredSparsity}, and dynamic in-training drop-and-grow algorithm \cite{evci2020rigging,frankle2019lottery}. Introducing sparsity in neural networks has been shown to improve model performance, accelerate convergence, and regularize the network improving model generalization \cite{hoefler2021sparsity}.

This study aims to reduce the computational cost of training Bayesian neural networks by decreasing the number of Bayesian parameters. We achieve this by:
\begin{enumerate}
    \item Promoting sparsity among Bayesian parameters based on strong deterministic predictive performance (Figure \ref{fig:sparse-wise}).
    \item Introducing a training method to integrate mixed deterministic and Bayesian parameters into a given network architecture.
    \item Demonstrating our novel training approach for medical image classification and segmentation, achieving competitive performance with over 95\% reduction in Bayesian parameters.    
\end{enumerate}

The proposed method, depicted in Figure \ref{fig:training}, initializes a partial Bayesian NN\footnote{Note we will use sparse and partial Bayesian interchangeably, as promoting sparsity in the Bayesian parameters renders the network \textit{partially} Bayesian.} using estimated points from a trained deterministic neural network (NN). Specific parameters are designated as Bayesian based on their saliency, with the degree of ``Bayesian-ness" controlled by a hyperparameter $r_{bayes}$. Experimental results in classification and segmentation demonstrate that assigning only 1\% of the network parameters as Bayesian can yield high-quality predictive uncertainty while preserving model performance. This approach accelerates BNN convergence with significant computational cost savings. 

\section{Method}
\subsection{Problem Setup}
Consider a dataset $\mathcal{D}=\{x_n, y_n\}_{n=1}^N$ where $x_n \in \mathbb{R}^{W\times H\times C}$, where $W$, $H$, $C$, represent the width, height, and number of channels for an input image, respectively. For segmentation $\{y^{(r)}_n \in \mathcal{Y}^{W\times H}\}$ represents the mask for rater $r$, for classification $\{y_n\in {1, ..., c}\}$ represents the categorical class $c$. The goal is to train a neural network to model the probabilistic distribution $p_\theta (y|x)$ where $\theta$ represents the network parameters. Let $\theta_d$ represent the deterministic point estimates, and $\theta_b$ represent Bayesian parameters, such that $\theta_b^{(i)}$ is parameterized by $\{\mu_i, \sigma_i\}$, the mean and standard deviation of the Gaussian distribution $\mathcal{N}_i(\mu_i, \sigma_i)$. The objective is to learn the probabilistic distribution of $p_\theta (y|x)$ where $\theta=\{\theta_d, \theta_b\}$ comprises a blend of deterministic and Bayesian parameters and reduce the number of Bayesian parameters to optimize model complexity and performance.

\begin{figure}[!ht]
\begin{center}
\captionsetup[sub]{font=small, margin={0pt,-10pt}}
    \begin{subfigure}{0.3\columnwidth}
    \begin{center}
        \includegraphics[trim={2cm 0 0 0},clip,height=0.8\linewidth]{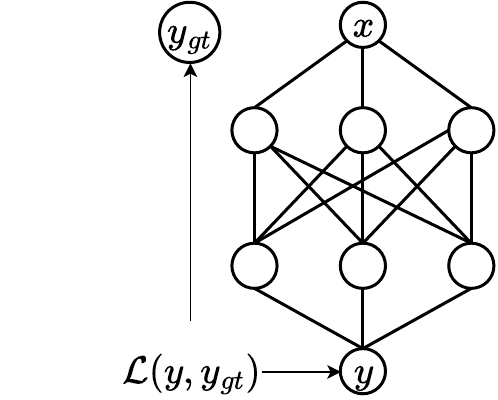}
        \caption*{\textbf{Step 1}: Deterministic}
    \end{center}
    \end{subfigure}
    \hspace{3mm}
    \begin{subfigure}{0.3\columnwidth}
    \begin{center}
        \includegraphics[trim={0 0 1.5cm 0},clip,height=0.8\linewidth]{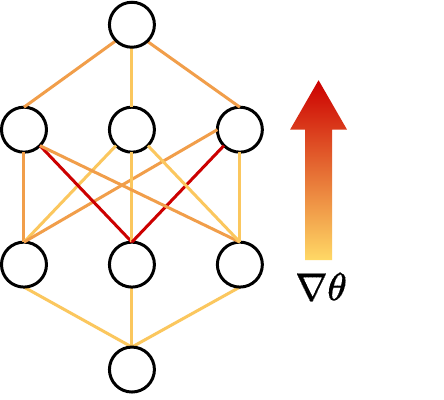}
        \caption*{\textbf{Step 2}: Sensitivity Analysis}
    \end{center}
    \end{subfigure}
    \hfill
    \begin{subfigure}{0.3\columnwidth}
    \begin{center}
        \includegraphics[trim={0 0 0.4cm 0},clip,height=0.8\linewidth]{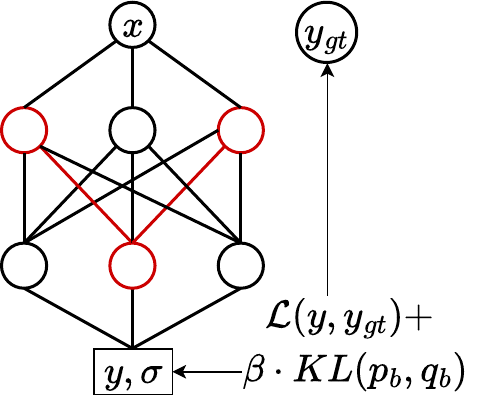}
        \caption*{\textbf{Step 3}: Sparse Bayes}
    \end{center}
    \end{subfigure}
    \caption{Our proposed training of sparse (partial) Bayesian network. \textbf{Step~1}: Train a deterministic model by minimizing the negative log likelihood $\mathcal{L}(y, y_{gt})$ where the parameters are represented as point estimates. \textbf{Step~2}: Perform a gradient-based sensitivity analysis, denoted as $\nabla \theta$, and identify the Topk connections corresponding to the highest gradients (in red). \textbf{Step~3}: Train a sparse (partial) Bayesian model with the Topk connections as Bayesian parameters and the remaining network as deterministic by minimizing the Evidence Lower Bound (ELBO) loss $\mathcal{L}(y, y_{gt})+\beta \cdot KL\left(p_b(\theta), q_b(\theta)\right)$, where $p_b(\theta)$ and $q_b(\theta)$ are the prior and posterior distributions for the $\theta_b$ Bayesian parameters.}
    \label{fig:training}
\end{center}
% \vskip -0.2in
\end{figure}

\subsection{Sparse Bayesian Networks}

We introduce a simple procedure to train a mixed-parameter model, regardless of the chosen architecture or application: (1) train a deterministic model, (2) compute parameter gradients to select the Topk most predictive parameters, (3) train a partial Bayesian model. The following is a detailed description of each step:

\subsubsection{(1) Train a deterministic model:} given a dataset $\mathcal{D}$, train a neural network $f_d(\cdot)$ by minimizing the negative log likelihood $\sum_i \mathcal{L}(f_{\theta_d}(x_i), y_i)$ (Figure \ref{fig:training}-Step 1). 

\subsubsection{(2) Sensitivity analysis:} Performing a gradient-based sensitivity analysis is seamlessly integrated into the backward pass through the network, thus incurring no additional computational overhead. Selecting the most salient parameters involves squaring the gradients. This process ensures that any large gradient, regardless of its sign ($+/-$), indicates a parameter with high predictive performance. Subsequently, select the Topk parameters, i.e. Topk$(\nabla \theta_i)^2$, where $k$ is the number of parameters designated as Bayesian. (see Figure \ref{fig:training}-Step 2). 

\subsubsection{(3) Train a partial Bayesian model:} A sparse masked-gradient approach is employed to have a mixture of deterministic and Bayesian parameters within the network layers. 

\paragraph{\textbf{Initialization:}} Firstly, point estimates from the deterministic model are set to the mean $\mu_i$ parameter values to initialize the partial Bayesian model. This ensures a robust starting point for optimizing uncertainty learning within the Bayesian network. Then, for each layer within the model, a mask is generated based on the Topk gradient values of its weights. These masks serve two main purposes: initializing the $\sigma_i$ parameter associated with Bayesian weights and facilitating sparse gradient updates. This enables the modeling of mixed-parameter type, i.e. deterministic and Bayesian, within a single layer. The deterministic weights do not contribute to the Kullback-Leibler (KL) divergence term. Consequently, deterministic weights are represented as $\delta_{i}(\mu_{i})$ delta functions, whereas Bayesian weights are modeled as $\mathcal{N}_{i}(\mu_{i}, \sigma_{i})$ distributions. The extent of ``Bayesian-ness" of the model is hyper-parameterized by an $r_{bayes}$ factor, which determines the rate (or percentage) of Bayesian parameters over the total number of parameters. 

\paragraph{\textbf{Training:}} The partial Bayesian model is trained with variational inference \cite{kingma2013VI} using the reparameterization trick \cite{kingma2015reparam}. Five posterior samples were used at train and inference time to reduce the computational overhead associated with larger samples. The network is trained by minimizing the evidence lower bound (ELBO) loss: 
\begin{equation}\label{eqn:elbo}
\mathbb{E}\left({\log p(x|\theta) + \beta \cdot KL\left(\mathcal{N}_{p}, \mathcal{N}_{q}\right)}\right) = \mathbb{E}\left(\log p(x|\theta) \right) + \beta \cdot \mathbb{E}\left(\log \frac{q(\theta_b)}{p(\theta_b)}\right) 
\end{equation}
where $\beta$ is the weight of the KL divergence term that minimizes the difference between the prior distribution $p(\theta_b)$ and the posterior $q(\theta_b)$. The prior distribution is the standard normal $\mathcal{N}(0,1)$. The first term in the ELBO loss is the expectation of the maximum likelihood estimation (see Figure \ref{fig:training}-Step 3). 

It is important to emphasize that the training procedure described above applies to any supervised training approach, regardless of the task. The experimental setup presented next aims to illustrate the effectiveness of our proposed training scheme for both medical image classification and segmentation tasks, enabling the estimation of uncertainty in both objectives.

\section{Experimental Results}

The objective of the following experimental setup is to showcase the versatility of the training paradigm in both medical image classification and segmentation tasks rather than achieving state-of-the-art performance. A standard deterministic model, a fully Bayesian model, and an ensemble approach are comparative baselines for task performance and uncertainty quantification. The proposed method is demonstrated on the classification of ChestMNIST dataset \cite{medmnistv2,wang2017chestx} and binary segmentation of LIDC-IDRI \cite{lidc1,lidc2,lidc3} and ISIC \cite{isic1,isic2} datasets. ChestMNIST contains 112,120 X-Ray images from 30,805 patients with 78,468/11,219/22,433 as a train/validation/test split with an input size of $224\times224$. LIDC-IDRI contains 1018 clinical thoracic CT scans, cropped into 15,096 patches of size $128\times128$, each with 4 manual segmentation masks; the data is divided into 4 bags for 4-fold train/test splits (0.75/0.25). ISIC dataset of skin lesions contains 2,594/100/1,000 train/validation/test images, all resized to~$224\times224\times3$.

\subsection{Evaluation Metrics \& Training Details}
The performance metrics are determined by the primary evaluation criteria specific to each dataset. For ChestMNIST classification, accuracy and the area under the receiver operating characteristic curve (AUC). The reported multi-label AUC score is the average of the binary AUC scores. For LIDC-IDRI segmentation, the Dice score is used to measure model performance and intersection over union (IoU) for ISIC dataset. The Brier score is used as a proper scoring rule for evaluating the accuracy of the output probabilities $\text{Brier Score} = \frac{1}{c} \sum_{i=1}^{c} (\delta^*_{c} - p(y = c\mid x^*))^2$ where $\delta^*_c = 1$ if $c=y^*$, and  $\delta^*_c = 0$ otherwise. Entropy of Expectation (EoE) is used to compute the total uncertainty $\text{EoE} = - p\left(y=c\mid x\right) \log\left(p\left(y=c\mid x\right)\right)$. Lastly, the Expected Calibration Error per \cite{guo17calibration} is used to evaluate model calibration. Floating point operations (FLOPs) are used to measure computational efficiency.

Deterministic and partial Bayesian models were trained for 50 epochs, with a fixed learning rate of 0.01, and a weight decay of $1\times10^{-5}$ for $L_2$ regularization with a stochastic gradient descent optimizer. Batch normalization was used in all models, with a batch size of 50 for classification and 10 for segmentation. The models were trained using a weighted cross-entropy loss to address class imbalances effectively. All ensemble models for comparison were executed with 5 ensemble members. For all datasets, we demonstrate the impact of increasing the ``Bayesian-ness" ($r_{bayes}$) on the performance and uncertainty estimation, given a fixed computational budget by varying $r_{bayes} = (1\%, 5\%, 10\%, 20\%, 40\%, 80\%)$. The partial and fully Bayesian models were trained with ELBO \ref{eqn:elbo} loss and a static weight for the KL term of $\beta = 0.01$ for classification and annealed for segmentation from 0.2 to 0.01 as a function of epochs $\beta_i = \beta_{target} - (\beta_{target}-\beta_{init})\times({epoch}_i/{epoch}_{total})$. 

%Models were trained with NVIDIA RTX A6000 GPUs. \href{https://github.com/sinenomine0/SPBNN}{Code repository} includes reproducibility and software library information. 

% The fully Bayesian models required extended epochs for training due to their slower convergence rate, 400 for ChestMNIST, 200 for LIDC-IDRI (for each kfold), and 150 epochs for ISIC dataset.

\begin{table}[!t]
\caption{Performance metrics comparing benchmark performances with partial Bayesian models using different $r_{bayes}$ values (1\%, 5\%, 20\%) for classification and segmentation results. Our partial 1\%-model matches the performance of a 5-member ensemble with 30\% fewer FLOPs. *Bayesian model FLOPs are averaged across three datasets due to variable epochs needed for satisfactory convergence.~(See supporting information for statistics and additional metrics.)}
\label{tab:performance}
\resizebox{\columnwidth}{!}{
\begin{tabular}{|l|lll|lll|lll|l|}
\hline
\multirow{2}{*}{} & \multicolumn{3}{c|}{\textbf{ChestMNIST}} & \multicolumn{3}{c|}{\textbf{LIDC-IDRI}} & \multicolumn{3}{c|}{\textbf{ISIC}} & \multicolumn{1}{c|}{\multirow{2}{*}{\textbf{FLOPs}}} \\ \cline{2-10}
 & \multicolumn{1}{l|}{\textbf{Accuracy} $\uparrow$} & \multicolumn{1}{l|}{\textbf{Brier} $\downarrow$} & \textbf{Entropy} $\downarrow$ & \multicolumn{1}{l|}{\textbf{Dice} $\uparrow$} & \multicolumn{1}{l|}{\textbf{Brier} $\downarrow$} & \textbf{Entropy} $\downarrow$ & \multicolumn{1}{l|}{\textbf{IoU} $\uparrow$} & \multicolumn{1}{l|}{\textbf{Brier} $\downarrow$} & \textbf{Entropy} $\downarrow$ & \multicolumn{1}{c|}{} \\ \hline
\textbf{Deterministic} & \multicolumn{1}{l|}{0.899} & \multicolumn{1}{l|}{0.098} & 0.493 & \multicolumn{1}{l|}{0.710} & \multicolumn{1}{l|}{0.075} & 0.007 & \multicolumn{1}{l|}{0.801} & \multicolumn{1}{l|}{0.110} & 0.151 & 1$\times$ \\ \hline
\textbf{Ensemble} & \multicolumn{1}{l|}{0.936} & \multicolumn{1}{l|}{0.053} & 1.162 & \multicolumn{1}{l|}{0.687} & \multicolumn{1}{l|}{0.075} & 0.007 & \multicolumn{1}{l|}{0.808} & \multicolumn{1}{l|}{0.109} & 0.154 & 5$\times$ \\ \hline
\textbf{Partial 1\%} & \multicolumn{1}{l|}{0.934} & \multicolumn{1}{l|}{0.064} & 0.367 & \multicolumn{1}{l|}{0.800} & \multicolumn{1}{l|}{0.004} & 0.010 & \multicolumn{1}{l|}{0.783} & \multicolumn{1}{l|}{0.068} & 0.154 & 3.36$\times$ \\ \hline
\textbf{Partial 5\%} & \multicolumn{1}{l|}{0.931} & \multicolumn{1}{l|}{0.067} & 0.404 & \multicolumn{1}{l|}{0.750} & \multicolumn{1}{l|}{0.005} & 0.016 & \multicolumn{1}{l|}{0.762} & \multicolumn{1}{l|}{0.076} & 0.197 & 3.45$\times$ \\ \hline
\textbf{Partial 20\%} & \multicolumn{1}{l|}{0.925} & \multicolumn{1}{l|}{0.083} & 1.303 & \multicolumn{1}{l|}{0.750} & \multicolumn{1}{l|}{0.013} & 0.050 & \multicolumn{1}{l|}{0.677} & \multicolumn{1}{l|}{0.110} & 0.300 & 3.8$\times$ \\ \hline
\textbf{Bayesian} & \multicolumn{1}{l|}{0.723} & \multicolumn{1}{l|}{0.215} & 3.794 & \multicolumn{1}{l|}{0.670} & \multicolumn{1}{l|}{0.110} & 0.300 & \multicolumn{1}{l|}{0.675} & \multicolumn{1}{l|}{0.111} & 0.273 & *15$\times$ \\ \hline
\end{tabular}%
}
\end{table}

\subsection{ChestMNIST Classification} 
We first evaluate a ResNet-18 \cite{he2016resnet} our method on multi-label classification of ChestMNIST dataset \cite{medmnistv2,wang2017chestx}. Figure \ref{fig:performance} summarizes the comparative performance of sparse Bayes with different $r_{bayes}$ values, with ensemble and fully Bayesian approaches. With increasing $r_{bayes}$ and a fixed computational budget, there's a decline in test performance metrics, both on classification error (Figure \ref{fig:test_error}) and uncertainty metrics (Figure \ref{fig:brier},\ref{fig:entropy}). Comparing the performance of a 1\% model to a 5-member ensemble, it's evident that the sparse model performs competitively, particularly in entropy. The ensemble yields am entropy of $1.162$, whereas the 1\% model achieves an entropy of $0.367$ (see Table \ref{tab:performance}), with $\sim80\%$ fewer parameters. The fully Bayesian model performs poorly, with a 20\% decline in accuracy and a $10\times$ entropy and $\sim5\times$ FLOPs of the 1\% partial model. For further details and metrics, please refer to the supporting information.

 \begin{figure*}[!b]
    \begin{subfigure}{0.47\columnwidth}
    \centering
        \includegraphics[trim={0.4cm 0.8cm 0.4cm 0.4cm},clip, width=1\linewidth]{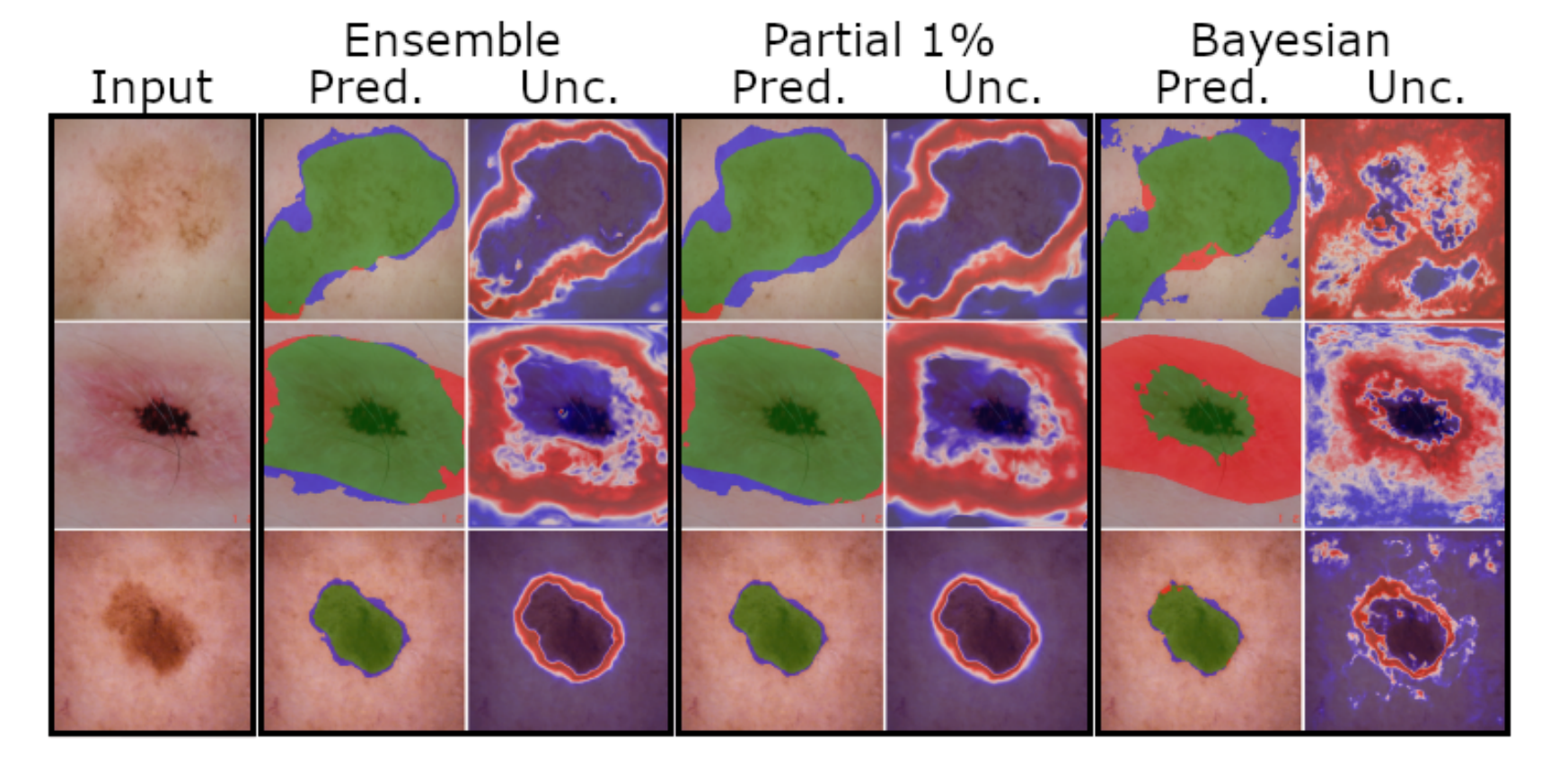}
        \caption{ISIC}\label{fig:samples-isic}
    \end{subfigure}
    \hfill
    \begin{subfigure}{0.55\columnwidth}
    \centering
        \includegraphics[trim={0cm 0.8cm 0cm 0cm},clip, width=1\linewidth]{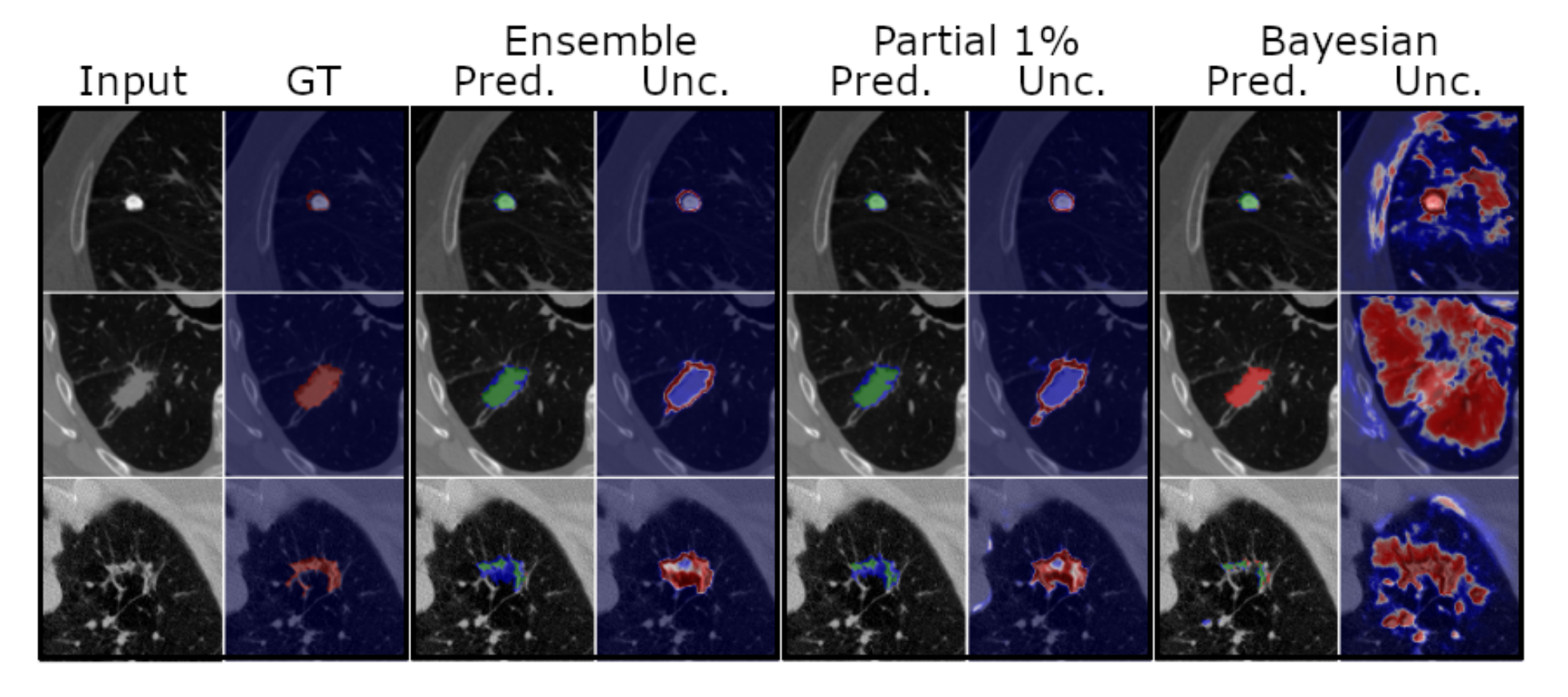}
        \caption{LIDC-IDRI}\label{fig:samples-lidc}
    \end{subfigure}
    \caption{Segmentation samples for 5-member ensemble, 1\% partial Bayesian, and fully Bayesian models with input image on the far left. Predictions mask overlays show true positive (green), false positive (blue), and false negative (red). The uncertainty map is the entropy of the output probability, showing regions of high (red) and low (blue) uncertainty. (b) LIDC-IDRI includes inter-rater variability (2nd column). Our partial 1\% is at par with ensembles at a lower cost. (Zoom in for a better view of the details.)}
    \label{fig:samples}
% \vskip -0.2in
\end{figure*}

\subsection{LIDC-IDRI \& ISIC Segmentation} 
\begin{figure*}[t!]
\begin{center}
    \begin{subfigure}{0.32\columnwidth}
    \centering
        \includegraphics[width=0.9\linewidth]{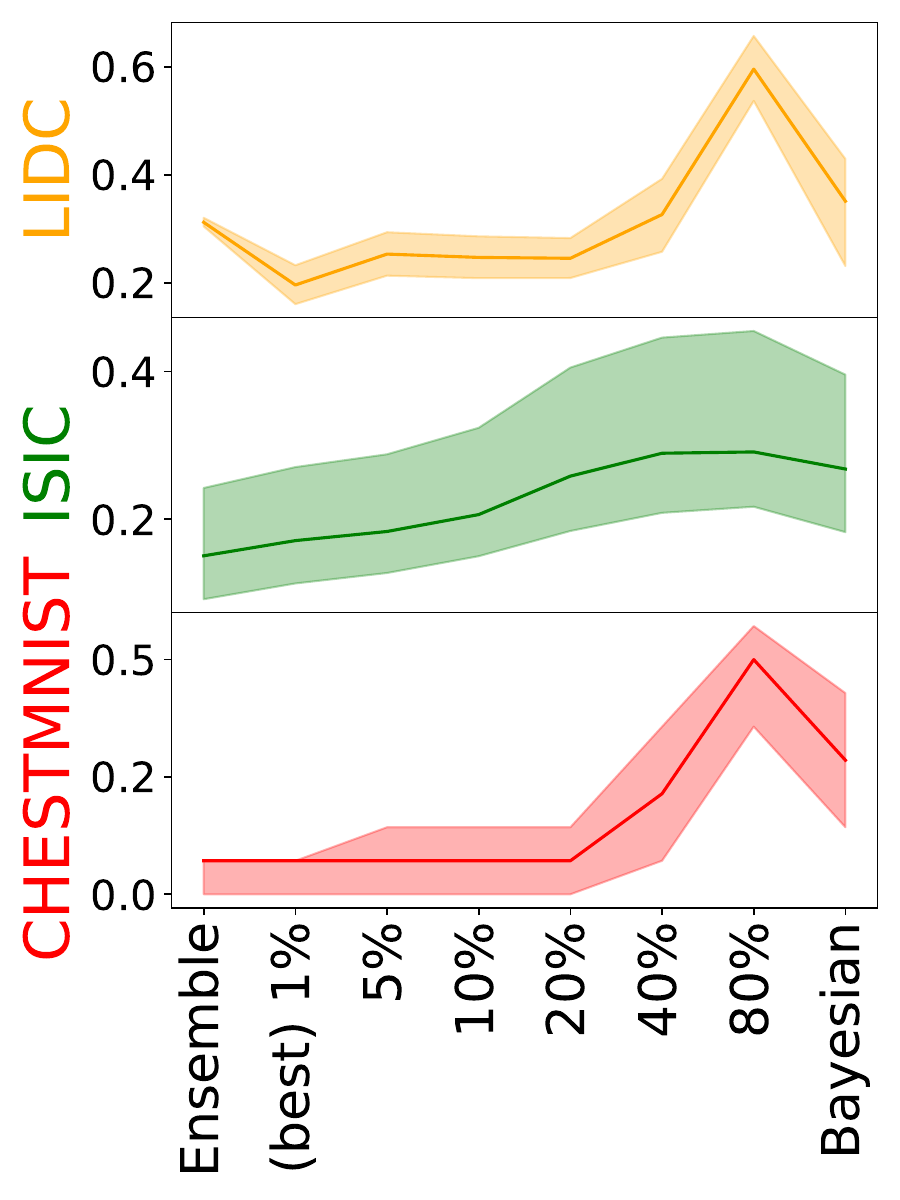}
        \caption{Test Error $\downarrow$}\label{fig:test_error}
    \end{subfigure}
    \hfill
    \begin{subfigure}{0.32\columnwidth}
    \centering
        \includegraphics[width=0.9\linewidth]{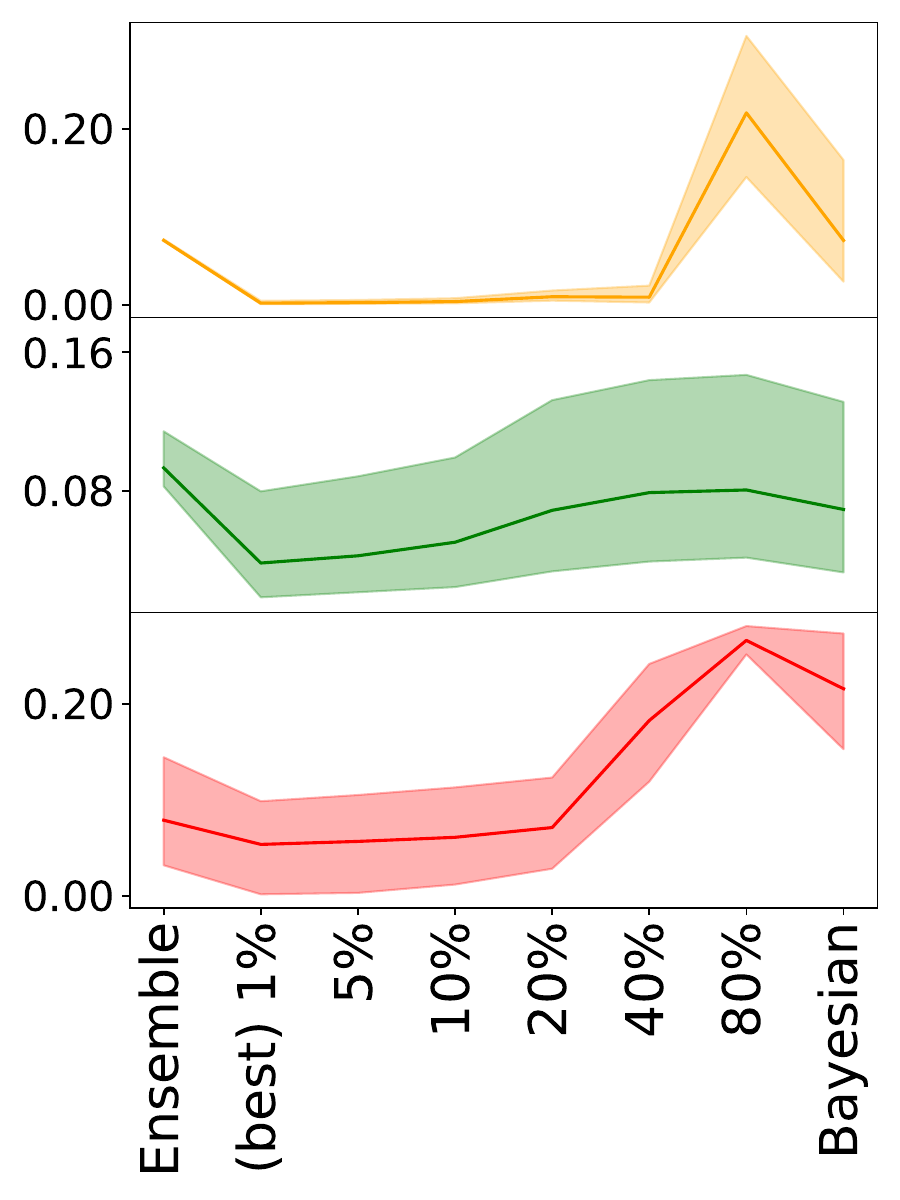}
        \caption{Brier Score $\downarrow$} \label{fig:brier}
    \end{subfigure}
    \hfill
    \begin{subfigure}{0.32\columnwidth}
    \centering
        \includegraphics[width=0.9\linewidth]{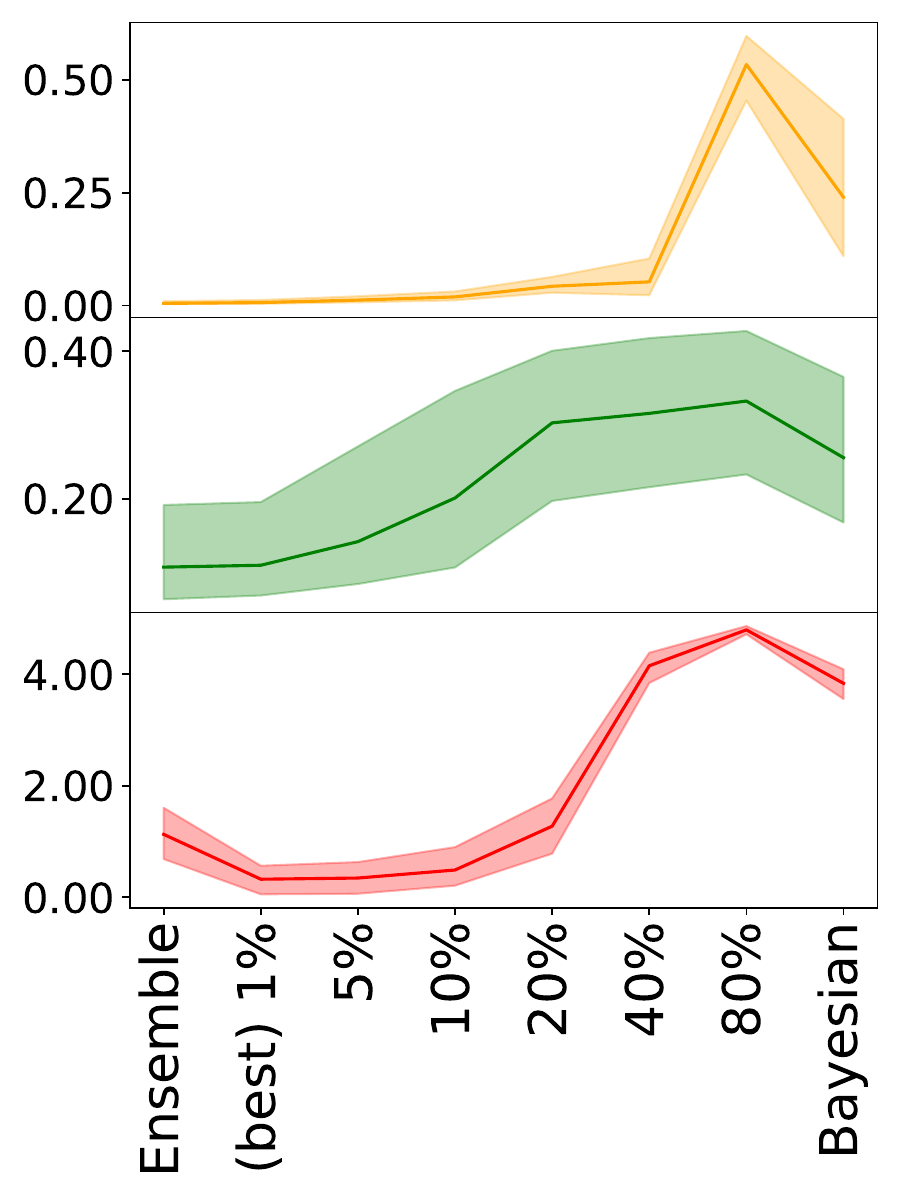}
        \caption{Entropy $\downarrow$}\label{fig:entropy}
    \end{subfigure}
    \hfill
    \caption{Performance comparison of partial Bayesian models with $r_{bayes}=$~(1\%, 5\%, 10\%, 20\%, 40\%, 80\%), 5-member ensembles, and fully Bayesian models for classification and segmentation tasks. Mid-line is the median metric value. Shaded area indicates the 25-75\% interquartile range. Test Error is computed as $(1-\text{accuracy/Dice/IoU})$.}
    \label{fig:performance}
\end{center}
% \vskip -0.25 in
\end{figure*}

Next, we evaluate a UNet\cite{ronneberger2015u} on the segmentation of LIDC-IDRI and ISIC datasets. For LIDC-IDRI, we randomly pick a mask of the 4-ground truth masks at each training step. Our experiments demonstrate the effectiveness of our proposed training paradigm with sparse Bayesian parameter representation, achieving competitive performance against benchmark Bayesian and Ensemble methods (Table \ref{tab:performance}, Figure \ref{fig:performance}). Setting only 1\% of the network's parameters to Bayesian yields comparable or superior performance to ensembles while requiring fewer floating point operations (FLOPs) by 30\% and significantly fewer parameters 80\% less than ensembles (Table \ref{tab:performance}). Comparing the 1\% partial Bayesian model to the fully Bayesian model (Figure \ref{fig:performance} and Figure \ref{fig:samples}), our training approach demonstrates significantly better performance with a fraction of the required FLOPs. For instance, in the segmentation of LIDC lung nodules, the Partial 1\% model achieves a Dice score of $0.80\pm0.01$ and an Entropy of $0.010\pm0.001$, while the fully Bayesian model achieves a Dice score of $0.67\pm0.09$ and an Entropy of $0.3\pm0.1$. Moreover, our approach incurs less than 20\% of the computational cost (FLOPs) of training the fully Bayesian model and approximately 50\% fewer parameters.

Qualitatively, Figure \ref{fig:samples} demonstrates the consistency of our model across both segmentation tasks. In particular, Figure \ref{fig:samples-lidc} illustrates how our approach aligns not only with the ground truth majority vote but also with the uncertainty of the ground truth segmentation. Despite low Bayesian rates (1\%), the partial Bayesian model effectively expresses uncertainty, particularly evident in regions with high uncertainty in LIDC-IDRI and ISIC datasets, consistent with areas of higher ambiguity (second example), such as near borders of ROIs or unclear boundaries of skin lesions (Figure \ref{fig:samples-isic}). Conversely, training a fully Bayesian network results in inferior performance, higher uncertainty, and higher false negative rate (demonstrated in the red overlay regions), with significantly increased computational costs due to slow model convergence (Figure \ref{fig:performance}).

\section{Conclusion}
In this paper, we present a training procedure for efficiently training sparse Bayesian networks, achieving competitive performance and predictive uncertainty estimation compared to fully Bayesian and ensemble methods. Our approach demonstrates significantly fewer parameters and lower computational requirements without compromising task performance. Specifically, a network with only 1\% Bayesian parameters matches or surpasses ensemble performance and consistently outperforms fully Bayesian networks with orders of magnitude fewer FLOPs. This enables more cost-effective predictive uncertainty quantification without compromising performance, thus facilitating uncertainty-guided decision-making in medical image analysis. A limitation of our current work is the absence of exploration into sub-percent Bayesian parameterization. Future research will consider setting parameters as Bayesian at lower percentages, specifically below 1\%, to further alleviate the computational burden.

\subsubsection{\ackname} We acknowledge the support of the Natural Sciences and Engineering Research Council of Canada (NSERC) for PGS-D award (No. 545704), NSERC Discovery grant (RGPIN-2020-06558), and  Canada Research Chair in Shape Analysis in Medical Imaging.

\bibliographystyle{splncs04}
\bibliography{ref}

\end{document}